%% file: main.tex
\AtBeginDocument{%
  \providecommand\BibTeX{{%
    \normalfont B\kern-0.5em{\scshape i\kern-0.25em b}\kern-0.8em\TeX}}}

\documentclass[sigconf,nonacm]{acmart}

\copyrightyear{2025}
\acmYear{2025}
\setcopyright{rightsretained}
\acmConference[SIGCSE TS 2025]{Proceedings of the 56th ACM Technical Symposium on Computer Science Education V. 1}{February 26-March 1, 2025}{Pittsburgh, PA, USA}
\acmBooktitle{Proceedings of the 56th ACM Technical Symposium on Computer Science Education V. 1 (SIGCSE TS 2025), February 26-March 1, 2025, Pittsburgh, PA, USA}
\acmDOI{10.1145/3641554.3701791}
\acmISBN{979-8-4007-0531-1/25/02}

\usepackage{comment}
\usepackage[utf8]{inputenc} 
\usepackage[T1]{fontenc}    
\usepackage{hyperref}       
\usepackage{url}            
\usepackage{booktabs}       
\usepackage{amsfonts}       
\usepackage{amsmath}        
\usepackage{nicefrac}       
\usepackage{microtype}      
\usepackage{xcolor}         
\usepackage{tikz}           
\usepackage{wasysym}

\usepackage{caption}
\usepackage{subcaption}

\usepackage{pdfpages}
\usepackage{comment}

\usepackage[listings]{tcolorbox}
\usepackage{tcolorbox}

\usepackage[disable]{todonotes}
\usepackage{fancyvrb}

\usepackage{tikz}
\usetikzlibrary{tikzmark, decorations.pathreplacing}

\newtcolorbox{example}[1]{colbacktitle=white,coltitle=black!75!white,size=small,fontupper=\small,title={#1}}

\newcommand*\smallcircled[1]{\tikz[baseline=(char.base)]{
            \node[shape=circle,draw,inner sep=0.8pt] (char) {#1};}}

\title[Evaluating Language Models for Generating and Judging Programming Feedback]{Evaluating Language Models for Generating and Judging Programming Feedback}

\author{Charles Koutcheme}
\affiliation{%
  \institution{Aalto University}
  \city{Espoo}
  \country{Finland}
}
\email{charles.koutcheme@aalto.fi}
\orcid{0000-0002-2272-2763}

\author{Nicola Dainese}
\affiliation{%
  \institution{Aalto University}
  \city{Espoo}
  \country{Finland}
}
\email{nicola.dainese@aalto.fi}
\orcid{0000-0001-9806-419X}

\author{Sami Sarsa}
\affiliation{%
  \institution{University of Jyväskylä}
  \city{Jyväskylä}
  \country{Finland}
}
\email{sami.j.sarsa@jyu.fi}
\orcid{0000-0002-7277-9282}

\author{Arto Hellas}
\affiliation{%
  \institution{Aalto University}
  \city{Espoo}
  \country{Finland}
}
\email{arto.hellas@aalto.fi}
\orcid{0000-0001-6502-209X}

\author{Juho Leinonen}
\affiliation{%
  \institution{Aalto University}
  \city{Espoo}
  \country{Finland}
}
\email{juho.2.leinonen@aalto.fi}
\orcid{0000-0001-6829-9449}

\author{Syed Ashraf}
\affiliation{%
  \institution{Aalto University}
  \city{Espoo}
  \country{Finland}
}
\email{syed.ashraf@aalto.fi}
\orcid{0009-0007-0513-4613}

\author{Paul Denny}
\affiliation{%
  \institution{The University of Auckland}
  \city{Auckland}
  \country{New Zealand}
}
\email{paul@cs.auckland.ac.nz}
\orcid{0000-0002-5150-9806}

\renewcommand{\shortauthors}{Charles Koutcheme et al.}

\begin{document}



\begin{abstract}
\input{sections/00_abstract}

\end{abstract}

\maketitle

\input{sections/01_introduction}

\input{sections/02_related_work}
\input{sections/03_methodology}
\input{sections/04_results}
\input{sections/05_discussion}

\input{sections/06_conclusions}
\input{sections/07_acknowledgements}

\balance
\bibliographystyle{ACM-Reference-Format}  
\bibliography{biblio/pure_ai,biblio/eduai,biblio/cer,biblio/software}

\end{document}

%% file: sections/00_abstract.tex
The emergence of large language models (LLMs) has transformed research and practice across a wide range of domains. Within the computing education research (CER) domain, LLMs have garnered significant attention, particularly in the context of learning programming. Much of the work on LLMs in CER, however, has focused on applying and evaluating proprietary models. In this article, we evaluate the efficiency of open-source LLMs in generating high-quality feedback for programming assignments and judging the quality of programming feedback, contrasting the results with proprietary models. Our evaluations on a dataset of students' submissions to introductory Python programming exercises suggest that state-of-the-art open-source LLMs are nearly on par with proprietary models in both generating and assessing programming feedback. Additionally, we demonstrate the efficiency of smaller LLMs in these tasks and highlight the wide range of LLMs accessible, even for free, to educators and practitioners.

%% file: sections/01_introduction.tex
\section{Introduction}

High-quality and timely feedback is essential for students in programming courses.  Some types of feedback, such as whether a program runs or passes a provided test suite, are readily available via simple automated means \cite{keuning2016towards, paiva2022automated}.  However, feedback on the causes of subtle programming errors and suggestions for resolving them can be difficult to produce \cite{keuning2018systematic}.  Especially in large classes, providing accurate and personalised explanations of bugs as feedback to students can be a manual and time-consuming task for educators, and yet essential for reducing frustration and aiding learning.  

The automated generation of human-like feedback has recently been made possible thanks to the accessibility of state-of-the-art generative AI tools, such as ChatGPT.  In particular, API access to powerful large language models (LLMs) has sparked the development of many programming feedback tools that are now being deployed in classrooms~\cite{prather2024instructors}.  These include tools for generating improved error messages \cite{leinonen2023using}, aiding real-time debugging \cite{kazemitabaar2023studying}, explaining code \cite{liffiton2023codehelp, leinonen2023comparing} and tailoring next-step hints \cite{roest2024next}.  Such systems have been successful not only in generating feedback, but also in assessing feedback quality, offering the potential for generating high-quality feedback through iterative improvement.

Despite the promise of LLM-based feedback generation and evaluation approaches, the vast majority of research and usage in computing education contexts has relied on proprietary models such as GPT-4.  This reliance on closed-source LLMs is concerning for several reasons.  It involves sending potentially sensitive data to third parties with no guarantees on how the data will be used, provides little insight into how the models are trained or what deliberate or inadvertent biases they may contain, and may come with unpredictable licensing expenses \cite{kukreja2024literature}.  Open-source LLMs, on the other hand, are freely accessible and open for modification and distribution, and have started to become viable alternatives.  Nevertheless, very few studies have explored their capabilities for providing or assessing programming feedback.  

In this work, we investigate the potential of open-source models to produce high-quality feedback, and assess the quality of feedback generated by other LLMs.  We focus on feedback that provides explanations of bugs or issues in student-written programs and outlines the steps to address these issues.  While prior research suggests that open-source language models are competitive alternatives to proprietary models for generating feedback, the extent to which they can serve as judges (validators) of such feedback remains unclear.  Using a publicly available benchmark dataset of student-written programs, we address the following two research questions:

\begin{description}
\item[RQ1] How do open- and closed-source models compare with respect to the quality of their generated bug explanations and suggested fixes?
\item[RQ2] To what extent can open- and closed-source models assess the quality of programming feedback generated by other models relative to expert human judgment?
\end{description}

To answer our first research question, we generate explanations of bugs and their corresponding fixes using five state-of-the-art open-source and three popular proprietary language models.  We manually evaluate this feedback using a custom rubric that includes the completeness and comprehensibility of the explanations and the accuracy of the suggested fixes.  To answer our second research question, we use these expert human-generated ground truth labels to evaluate the performance of the models on the task of judging the programming feedback.  

Our findings suggest that open-source language models are competitive with proprietary models for both generating and assessing programming feedback.  Given the potential benefits of open-source models in terms of transparency, trust, and cost, we argue that they should be increasingly adopted in computing education contexts.

%% file: sections/02_related_work.tex
\section{Related Work}

\subsection{Using Language Models For Feedback}

Automating assessment of programming exercises and providing feedback on the exercises have been studied for decades within the computing education research domain~\cite{keuning2018systematic,messer2024automated,paiva2022automated}. Classically, much of the existing work on automating feedback has focused on informing students about mistakes in their code, while providing formative feedback has been less common~\cite{keuning2018systematic}. Providing suggestions on the location of the issue or hints on how to fix the issue can improve students' performance over just pointing out that a test failed~\cite{hao2022towards}, but manually creating quality feedback can be very time-consuming.

The recent emergence of powerful language models has led to researchers exploring their capabilities for programming feedback ~\cite{balse2023investigating,liffiton2023codehelp,hellas2023exploring,kiesler2023exploring,pankiewicz2023large,leinonen2023using,phung2023generating,edwards2024opportunities} and, in general, the observations on the quality or utility of feedback has evolved with the introduction of better language models~\cite{hellas2023exploring}. As an example, GPT-3 had high variability in the quality of feedback, at times generating incorrect and inconsistent feedback~\cite{balse2023investigating}, while GPT-3.5 would often provide meaningful feedback and find issues in code, but also often hallucinate issues that were not present in the code~\cite{hellas2023exploring}. Language models are also better at detecting some types of errors than others~\cite{hellas2023exploring,kiesler2023exploring}, being useful, especially for providing feedback on syntax or compilation errors~\cite{kiesler2023exploring,leinonen2023using,phung2023generating}. Despite the advances, even the state-of-the-art models like GPT-4 are still not on par with humans when generating feedback for programming exercises~\cite{phung2023generative}. At the same time, there are increasing amounts of evidence that language model-powered feedback systems and chatbots in programming~\cite{liffiton2023codehelp,hicke2023aita,hellas2024experiences,wang2024large} can aid students, at least when the programming languages or used frameworks are not brand new~\cite{hellas2024experiences}. 

Most existing work on language models for programming feedback in the computing education research context has focused on utilizing proprietary models (mainly from OpenAI). In contrast, the use of open-source models has received only little attention. Calls for increasing the use of open-source models have been voiced~\cite{yan2023practical}, already due to potential privacy issues related to sharing student data with language model providers. Work on utilizing open-source models for the task is also starting to emerge \cite{kotalwar2024hintsinbrowserbenchmarkinglanguagemodels,hicke2023aita,liu_2024_can}, where one of the research aspects has been contrasting the performance of open-source models to the proprietary ones; researchers have already observed that open-source models are on par with models such as GPT-3.5-Turbo for programming feedback~\cite{koutcheme_2024_open}.

In our work, our first research question re-investigates how various language models, including open-source ones, perform in explaining issues in student programs and providing fixes, complementing prior studies.

\subsection{Using Language Models as Judges}
\label{bg:llm-judge}

The idea of using an LLM to judge the output of other LLMs -- LLMs-as-judges -- was first studied in the work of Zheng et al. \cite{zheng2023judging}, showing good promise, but also limitations, e.g., in grading math and reasoning tasks. Since then, GPT-4 has been used in multiple studies as a judge of the quality of other LLMs' generations \cite{liu-etal-2023-g}, also in educational contexts \cite{koutcheme_2024_open,hicke2023aita}. 
Moreover, the reliance on GPT-4, a proprietary model, has sparked interest in leveraging other open-source language models to act as judges \cite{dubey2024llama3herdmodels}. 
Yet, recent work has highlighted the limitations of relying on a single language model for evaluating the quality of other language models' outputs, and suggested employing a diverse ensemble of smaller models from different LLM families as a jury for cheaper and less biased evaluations  \cite{verga2024replacingjudgesjuriesevaluating}.
When answering our second research question, we test this hypothesis by comparing the usage of single judges (both open-source and proprietary) and that of a jury of smaller open-source language models. 

%% file: sections/03_methodology.tex
\section{Methodology}

In this section, we describe our methodology for answering our two research questions. We first introduce the dataset used in our evaluations, then our methods used for answering RQ1 and RQ2. 

\subsection{Dataset}
\label{sec:data}

We use data from the Socratic guidance benchmark \cite{socratic}, which consists of 57 introductory-level programming assignments requiring students to write functions. Each of the assignments is accompanied by the associated test cases, a unique incorrect student solution, the ground truth description of a single bug in the program, a list of potential bug fixes, and several conversation threads between a fake student and a teaching assistant. 
The ultimate goal for the benchmark is evaluating LLMs' ability to help students using the socratic method, i.e., guiding students in finding a solution on their own, by asking a series of relevant questions that help their reasoning.

However, for this study, we focus solely on identifying the issues in the code and any required fixes, as it is a fundamental step for effective use of LLMs for socratic guidance -- models incapable of identifying issues in students' programs would be likely to provide them erroneous guidance. We leave the problem of evaluating LLMs for socratic guidance for future work. 

\subsection{Generating High-Quality Feedback}
\label{meth:generating}

Given a student's incorrect program, our goal regarding RQ1 is to evaluate LLMs' ability to provide two particular types of feedback: 
\textbf{an explanation of the bugs} in the student's program and \textbf{suggested fixes} for the found bugs.

\paragraph{Feedback Generation.}
We prompt the models to provide feedback according to the following example:

\vspace{0.2cm}
\input{data/prompts/feedback_prompt}
\vspace{0.2cm}

To elaborate, we provide a language model: $\smallcircled{0}$ a system prompt and $\smallcircled{1}$ a description of the task (with all the necessary contextual information), which results in output $\smallcircled{2}$. 

\paragraph{Feedback Language Models.}
We consider the following open-source models: Gemma-2B \cite{google_gemma}, Phi-3-mini \cite{abdin2024phi3} (3.8B parameters), Mistral-7B \cite{jiang2023mistral}, Llama-3.1.1-8B \cite{dubey2024llama3herdmodels}, Llama-3.1.1-70B \cite{dubey2024llama3herdmodels}. We chose these models because of their extensive documentation, community adoption, strong performance on code and language reasoning benchmarks (e.g., MMLU), their parameter count, and their ability to follow instructions. This selection covers the recent state-of-the-art models from various companies across the most used model sizes for LLMs. 
Furthermore, we also evaluate three of OpenAI's proprietary flagship models, GPT-3.5-turbo, GPT-4o-mini, and GPT-4o, which represent the current industry standards.

We query proprietary models using the OpenAI Python library, and open-source ones with HuggingFace \textsc{Transformers} Python library to simplify querying them through the HuggingFace Inference API. All models are evaluated using greedy decoding \cite{koutcheme_2024_open}.
Next, we explain the annotation process before detailing the grading rubric.

\paragraph{Annotation.} 
We use the eight models presented above, and the 57 programs of the benchmark, which results in 8 \texttimes ~57 = 456 model outputs. To answer our first research question, two annotators (two paper authors who are expert Python programmers) annotated all 456 model outputs. First, we selected 11 problems out of the 57 available problems using the manual annotation subset presented in \cite{socratic}. Then, the two annotators independently annotated 8 \texttimes ~11 = 88 model outputs with an initial description of each grading criterion on this subset. We then computed an inter-annotator agreement score using Cohen's Kappa coefficient. The resulting annotation process yielded a moderate inter-rater agreement of 0.54. After comparing annotations and resolving conflicts, the remaining feedback examples were split equally between the two annotators.
The final annotated dataset formed the basis for evaluating the quality of the feedback generated by the language models.

\paragraph{Grading Criteria.}
During the final annotation phase, each expert used the following grading criteria for evaluating the quality of a single generated bug explanation (\textsc{E}), and the quality of the generated fixes (\textsc{F}):


\begin{itemize}
    \item \textsc{EA - Explanation Accurate}: the explanation identifies and correctly explains the bug in the student program.
    \item \textsc{ES - Explanation Selective}: the explanation does not mention non-existent (or non-relevant) bugs or issues. 
    \item \textsc{EC - Explanation Clear}: the explanation is easy to understand for a novice programmer, presented in a readable format and contains the right amount of information. Note: this criterion is independent of the correctness of the explanations.
    \item \textsc{FA - Fixes Accurate}: the required bug fixes are laid out and explained.
    \item \textsc{FS - Fixes Selective}: no unnecessary or irrelevant changes are outlined;
    \item \textsc{FC - Fixes Clear}: the proposed fixes are succinct and mention the unique changes to perform in the code. 
\end{itemize}

These criteria extend prior work \cite{koutcheme_2024_open,hellas2023exploring,phung2023automating}. Criteria EA, ES, FA, and FS (resp. EC, and FC) represent how correct (resp. how understandable) the explanations and criteria are. 
The annotators used the following guidelines: for each ground truth bug (provided in the dataset), match the bug with one (or several) model-generated explanations. If any generated model bug descriptions did not match the ground truth, we set the criteria ES to false, unless the model explicitly noted the irrelevance of the unmatched bug. Then, independently of the correctness, we looked at whether or not a novice programmer unaware of the real issues could understand the meaning of the provided bug description. We follow the same strategy for the fixes. Moreover, for the clarity criterion, we ensure that the fixes provide clear descriptions of changes with snippets or at least highlight changes in a repaired program (if present).

\subsection{Automatic Feedback Evaluation}

In this subsection, we present the methods we used to automatically evaluate the quality of LLM-generated feedback using other language models (answering RQ2). We explored two approaches: a single LLM as a judge, and an ensemble of LLMs as a jury on two scenarios, depending on whether a reference answer is available or not. We first describe how we generate the responses to the grading criteria using a single LLM as a judge for the two scenarios. We then outline our ensemble of LLMs and how we obtain the jury annotations.

\paragraph{No Reference Answer Available.}
Given the feedback generated by a language model, we prompt another language model (the judge) to grade this feedback (according to the criteria outlined in Section~\ref{meth:generating}) using the prompting strategy shown and described below: \\

\input{data/prompts/sag_grading_prompt}

Before asking the judge language model to grade the quality of the feedback $\smallcircled{3}$ (using the grading criteria list as part of the prompt), we first ask the judge to generate its own descriptions of the bugs and fixes in the student program, as described in Section~\ref{meth:generating} ($\smallcircled{0}$$\smallcircled{1}$$\smallcircled{2}$).
This strategy is a form of zero-shot-chain-of-thought 
Single Answer Grading  (SAG) \cite{zheng2023judging}, i.e., the judge uses its solution to the ``problem'' as a reasoning step to grade the solution of another language model.
This first scenario without a reference answer is applicable for educators and practitioners interested in evaluating language models' feedback abilities on their private datasets without the need for ground truth annotations.

\paragraph{Reference Answer Available.}
Educators may also be interested in evaluating their language models on existing benchmarks containing ground truth annotations of issues. Although such benchmarks are not abundant, 
we expect more to come as educational AI advances and becomes more widespread. In consideration of this scenario, we experiment with providing the judge with the ground truth descriptions of issues, instead of the more error-prone approach of generating them using the judge itself. The example below illustrates our prompting strategy:

\vspace{0.2cm}
\input{data/prompts/gag_grading_prompt}

\vspace{0.1cm}

This prompting strategy is a form of reference grading \cite{zheng2023judging}. We refer to this strategy as GAG (Ground truth Annotated Grading). We note that for both strategies (SAG, GAG), we extract the response to each grading criterion from the final judge model output.


\paragraph{Ensemble of Judges.} 
Prior work suggests that using a single LLM as a judge has limitations. For example, GPT-4 favours outputs from OpenAI's GPT line of models~\cite{rajani2023llm_labels}. As mentioned in Section \ref{bg:llm-judge}, instead of using a single LLM, Verga et al.~\cite{verga2024replacingjudgesjuriesevaluating} showed that using multiple language models from different families can address many of these issues. We aim to test their hypothesis in our educational context. We prompt three popular open-source language models to provide their judgement, and then, we combine the model decisions using majority voting separately for each criterion. For example, given three LLMs outputting `yes', `yes', and `no' respectively for a given criterion, the final ensemble result will be `yes'. We consider this ``jury'' both when reference answers are available (GAG) and not available (SAG). 

\paragraph{Judge and Jury Language Models.}
We evaluate both proprietary and open-source models as judges. We use GPT-3.5-turbo and extend prior work \cite{koutcheme_2024_open} by also including GPT-4o, and GPT-4o-mini. We compare these proprietary models against state-of-the-art open-source language models Phi-3-mini, Llama-3.1-8B, Mistral-7B, and Llama-3.1-70B. 
For the jury, we use the three language models: Llama-3.1-8B, Mistral-7B, and Phi-3-mini. We selected these three strong LLMs due to them being at the top of the leaderboards for LLMs of their size, and being from different families of models. 
As with the feedback generation, we query these language models using the OpenAI and Huggingface \textsc{Transformers} Python libraries and obtain the outputs using greedy decoding.

\paragraph{Evaluation}

To answer RQ2, we compare the evaluation of each judge/jury model for the 456 generated feedback outputs against the manually written ground truth annotations, with respect to our grading criteria (see Section \ref{meth:generating}). We then report the performance of each judge/jury across all of their outputs and for our two scenarios. We report the weighted average version of the $f_{0.5}$ score as in \cite{koutcheme_2024_open} (although they used plain $f_{0.5}$) for each judge/jury, as this metric accounts for false positives and potential class imbalances. We also report the kappa score to complement our observations. 

%% file: data/prompts/feedback_prompt.tex
\noindent
\begin{minipage}{\linewidth}
\centering
\begin{tcolorbox}[fontupper=\scriptsize, 
boxsep=0.8mm, left=0.8mm, bottom=0.8mm, top=0.8mm,
right=3.5mm, right skip=1cm, boxrule=0.1mm,
arc=4mm, arc is curved, 
rounded corners, sharp corners=northwest,
before=\smallcircled{0}
]

You are a CS professor teaching introductory programming using Python.

\end{tcolorbox}
\end{minipage}%
\hfill
\begin{minipage}{\linewidth}
\centering
\begin{tcolorbox}[fontupper=\scriptsize, 
boxsep=1mm, left=0.8mm, bottom=0.8mm, top=0.8mm,
right=3.5mm, right skip=1cm, boxrule=0.1mm,
arc=4mm, arc is curved, 
rounded corners, sharp corners=northwest,
before=\smallcircled{1}
]
Below are a problem description, test cases, and an incorrect program written by a student (i.e., it does not pass all test cases). \\ 

\vspace{-0.1cm}

\textcolor{blue}{<problem description>, <test cases>, <student code>} \\ 

\vspace{-0.1cm}

Your tasks are as follows: \\
    
1. **Explain Bugs**: List and explain all the bugs in the program that prevent it from passing the unit tests. Please explain each bug in 1-2 sentences. Do not suggest performance improvements. \\

2. **Provide Fixes**: For each bug, suggest a code fix by describing the change in a concise sentence. You can specify a replacement, insertion, deletion, or modification of one or several lines of code. 
\\

Please focus on providing clear and concise explanations and fixes, 
and avoid suggesting unnecessary changes to the code.

\end{tcolorbox}
\end{minipage}%
\hfill
\begin{minipage}{\linewidth}
\centering
\begin{tcolorbox}[fontupper=\scriptsize,
boxsep=1mm, left=1mm, right=1mm, bottom=1mm, top=1mm,
arc is curved, arc=3mm,
rounded corners, sharp corners=northeast,
halign=flush right, left skip=4.5cm, width=3.5cm,
colback=red!5!white,
after=\smallcircled{2},
boxrule=0.1mm,
]

List of bugs and fixes
\end{tcolorbox}
\end{minipage}

%% file: data/prompts/sag_grading_prompt.tex
\begin{minipage}{\linewidth}
\centering
\begin{tcolorbox}[fontupper=\scriptsize, 
boxsep=0.1mm, right skip=4cm, left skip=3.5cm, boxrule=0.1mm, right=1cm
]

{...}


\end{tcolorbox}
\end{minipage}%
\hfill
\begin{minipage}{\linewidth}
\centering
\begin{tcolorbox}[fontupper=\scriptsize,
boxsep=1mm, left=1mm, right=1mm, bottom=1mm, top=1mm,
arc is curved, arc=3mm,
rounded corners, sharp corners=northeast,
halign=flush right, left skip=3.5cm, width=4cm,
colback=red!5!white,
after=\smallcircled{2},
boxrule=0.1mm,
]

List of judge-generated bugs and fixes
\end{tcolorbox}
\end{minipage}
\hfill
\begin{minipage}{\linewidth}
\begin{tcolorbox}[fontupper=\scriptsize, 
boxsep=1mm, left=1mm, bottom=1mm, top=1mm,
right=3.5mm, right skip=1cm, boxrule=0.1mm,
arc=4mm, arc is curved, 
rounded corners, sharp corners=northwest,
before=\smallcircled{3}
]

Below is a list of bugs and their fixes written by a teaching assistant. \\

\vspace{-0.15cm}

\textcolor{blue}{<List of bugs and fixes>} \\

\vspace{-0.15cm}

Your task is to evaluate the quality of the TA's feedback according 
to the grading criteria outlined below, using your own response as baseline feedback. \\ 

\vspace{-0.15cm}

\textcolor{blue}{<grading criteria>}
\textcolor{blue}{<grading guideline>}\\

\vspace{-0.15cm}

This evaluation will be conducted in two parts: \\ 

\vspace{-0.15cm}

1. Comparison: Compare the TA's feedback with your feedback.
Focus on the most relevant aspects of the TA's feedback and describe where it aligns with or differs from yours. Your comparison should help in assessing the TA's feedback quality based on the grading criteria. \\ 

\vspace{-0.15cm}

2. Grading List: Conclude with each criterion listed on a separate line in the following format:
"criteria: yes/no"

\end{tcolorbox}
\end{minipage}%
\hfill
\begin{minipage}{\columnwidth}
\centering
\begin{tcolorbox}[fontupper=\scriptsize,
boxsep=1mm, left=1mm, right=1mm, bottom=1mm, top=1mm, boxrule=0.1mm,
arc is curved, arc=3mm,
rounded corners, sharp corners=northeast,
halign=flush right, left skip=4.5cm, width=3cm,
after=\smallcircled{4},
colback=red!5!white]
... \\ 
\textcolor{red}{<FS - yes/no>} \\ 
\end{tcolorbox}
\end{minipage}

%% file: data/prompts/gag_grading_prompt.tex
\noindent
\begin{minipage}{\linewidth}
\centering
\begin{tcolorbox}[fontupper=\scriptsize, 
boxsep=0.8mm, left=0.8mm, bottom=0.8mm, top=0.8mm,
right=3.5mm, right skip=1cm, boxrule=0.1mm,
arc=2mm, arc is curved, 
rounded corners, sharp corners=northwest,
]

You are a CS professor teaching introductory programming using Python.

\end{tcolorbox}
\end{minipage}%
\hfill
\begin{minipage}{\linewidth}
\centering
\begin{tcolorbox}[fontupper=\scriptsize, 
boxsep=1mm, left=0.8mm, bottom=0.8mm, top=0.8mm,
right=3.5mm, right skip=1cm, boxrule=0.1mm,
arc=4mm, arc is curved, 
rounded corners, sharp corners=northwest,
]
Below are a problem description, test cases, and an incorrect program written by a student (i.e., it does not pass all test cases). You are also provided with the ground truth description of a bug in that program and the required fixes. \\ 

\vspace{-0.15cm}

\textcolor{blue}{<description>, <test cases>, <code>, <bug description>, <bug fixes>} \\ 

\vspace{-0.15cm}

Below is a list of bugs and their fixes written by a teaching assistant. \\

\vspace{-0.15cm}

\textcolor{blue}{<LLM feedback generation>} \\

\vspace{-0.15cm}

Your task is to evaluate the quality of the TA's feedback according 
to the grading criteria outlined below, using the provided ground truth bug description and fixes as reference.

\vspace{-0.15cm}

\Large{\textbf{...}}

\end{tcolorbox}

\end{minipage}%

%% file: sections/04_results.tex
\section{Results}

\subsection{Generating Feedback}

Table \ref{table:feedgen} shows the performance of each language model on various grading criteria, including both individual and grouped criteria, based on human evaluations. We make the following observations.

\input{data/table/feedback}

\paragraph{Open-Source vs. Proprietary Models.}
The GPT-4o and GPT-4o-mini models show very strong performance across nearly all individual and grouped criteria, beating all other models. Among the open-source models, there is significant variance in performance across different criteria. For instance, Llama-3.1-70B performs on par with or better than GPT-3.5-turbo on the accuracy and clarity of explanations (EA, EC), and the clarity of fixes (FC). Llama-3.1-70B remains also competitive with both GPT-4o-mini and GPT-4o for generating perfect explanations ($E_{all}$), and perfect fixes ($F_{all}$). In contrast, smaller models such as Gemma-2b perform poorly across the board. However, size alone does not determine performance, as other small open-source language models, like Phi-3-mini (with 3.8 billion parameters -- slightly more parameters than Gemma-2b), despite their smaller size, perform decently well on several criteria, notably for generating accurate explanations and fixes (EA, FA). 

\paragraph{Strengths and Weaknesses.}
Each model has its strengths and weaknesses. However, we notice that most models struggle with selectivity (i.e., they identify irrelevant issues or fixes), while they generally produce clear outputs (i.e., well-formatted and understandable responses). When looking at the feedback generations, the stronger models (e.g. Llama-3.1-70B, and the GPTs) often added performance suggestions (e.g. replace a for loop with a built-in function), while the other models often added incorrect outputs. This indicates a broader challenge in developing models that can effectively identify and focus on relevant issues without including redundant or irrelevant information. Improvements in this area could lead to substantial overall performance gains.

\paragraph{Explanation, Fixes, and Repairs.}
We can observe a direct relationship between models' abilities to explain issues and their ability to generate fixes: overall, language models that produce more accurate (resp., more selective) explanations tend to generate more accurate (resp., more selective) fixes. During our experiments, we also noticed that the generated feedback often included repaired programs after the fixes, even though we did not prompt the models for this.
Although program repairs are not our primary focus, they represent another valuable form of feedback for students and can later be leveraged for hint generation \cite{kotalwar2024hintsinbrowserbenchmarkinglanguagemodels}. Building on this, we hypothesize that a language model's performance in generating accurate and selective fixes provides insights into its ability to generate high-quality program repairs—repairs that are functionally correct and preserve the intent of the student’s original code \cite{koutcheme2024benchmarkingeducationalprogramrepair}.
If our hypothesis holds, our findings complement prior work \cite{koutcheme-etal-2024-using} suggesting that repair abilities may serve as a proxy for explanation quality in language models. This connection could enable educators to more easily identify language models suited for generating effective programming feedback.

\subsection{Evaluating Feedback}

Table \ref{tab:judge_results} shows the results of the judgment task, detailing the $f_{0.5}$ scores and kappa scores for each language model under the two scenarios (SAG and GAG). We can make several observations from these results:

\input{data/table/judge}

\paragraph{Open-Source vs. Proprietary Models.}
Without ground truth bug descriptions and fixes, LLama-3.1-70B shows superior judging ability compared to GPT-3.5-turbo and smaller open-source models. The model performs comparably to GPT-4o-mini and GPT-4o across many individual criteria. It even surpasses OpenAI's flagship models in average $f_{0.5}$-scores overall criteria (i.e., average scores across self-generated and other-generated feedback). 
While larger models generally perform better, smaller models such as Phi-3-mini and Mistral also demonstrate strong judging performance, rivalling GPT-3.5-turbo, a previous state-of-the-art model. These small models also rival top language models in terms of explanation accuracy (EA), explanation clarity (EC), and fixes clarity (FS). However, they still lag in terms of judging selectivity. 
Providing models with ground truth descriptions of bugs and required fixes (GAG setting) leads to significant performance gains (nearly 10\%) over relying solely on their explanations (SAG setting). This improvement is most pronounced for models like Mistral, which struggle with selectivity in the SAG setting. In the GAG setting, the performance gap between top models (LLama-3.1-70B, GPT-4o-mini, and GPT-4o) becomes negligible across all criteria, indicating these models are viable alternatives to one another.
However, in the GAG setting, Phi-3-mini's performance does not improve. When looking at the model reasoning process in that setting, the model appears to overly rely on exact matches with the ground truth, showing less flexibility in its evaluation. This behaviour might stem from the model's smaller size, which could limit its capacity for nuanced reasoning.

\paragraph{Kappa Scores.}
The low kappa scores in the SAG setting indicate that most model results could be due to random chance. When investigating the reasons for the scores, we see that most models tend to be overly positive, predicting `yes' the majority of the time. This phenomenon aligns with observations in \cite{koutcheme_2024_open}, highlighting a tendency of models to overestimate the quality of feedback. When provided with ground truth annotations, the results improve to reach a moderate level of agreement.

\paragraph{Self-Evaluation vs. Evaluation of Others.} 
Models perform better when evaluating other models' outputs than their own, with Llama-3.1-70b performing the best. This might be due to a bias toward positive evaluations, especially within models of the same family, as noted in previous research ~\cite{zheng2023judging}. Our result thus suggests that language models from different families should be used for generating and validating feedback \cite{phung2023automating}.

\paragraph{Ensemble Performance.} 
Combining multiple models into an ensemble does not improve judgment quality; instead, it biases the results. The ensemble approach, which combined the outputs of the models Phi-3-mini, Mistral-7B, and LLama-3.1-8B, did not yield better performance. This contrasts with previous work by Verga et al. \cite{verga2024replacingjudgesjuriesevaluating}, possibly due to the absence of few-shot examples, which provide the model with additional context and training data that could enhance the performance of the individual models.
In the previous study, models competitive with GPT-3.5-turbo were used, leading to better ensemble performance. Moreover, our method of taking the mode of the generations from three LLMs means that if two weaker models consistently disagree with the stronger model, the output can be negatively biased. This highlights a key limitation in ensemble approaches: the quality of the ensemble is highly dependent on the individual models' performance and their ability to complement each other.

%% file: data/table/feedback.tex
\begin{table}[htbp]
    \centering
    \caption{Feedback results for various language models based on human evaluations. Legend (see also grading criteria in \ref{meth:generating}): $E_{all}$ (resp. $F_{all}$): all explanations (resp. all fixes) related criteria are correct, ALL: all criteria are correct.
    }
    \resizebox{\linewidth}{!}{
    \input{data/table/tabular/feedback}
}
\label{table:feedgen}
\end{table}

%% file: data/table/tabular/feedback.tex
\begin{tabular}{l|rrrrrr|rrr}
    model & EA & ES & EC & FA & FS & FC & $E_{all}$ & $F_{all}$ & ALL \\
    \midrule
    Gemma-2b & 0.44 & 0.02 & 0.65 & 0.42 & 0.02 & 0.65 & 0.00 & 0.02 & 0.00 \\
    Phi-3-mini & 0.72 & 0.19 & 0.93 & 0.74 & 0.21 & 0.89 & 0.18 & 0.19 & 0.18 \\
    Mistral-7b & 0.70 & 0.23 & \textbf{0.96} & 0.70 & 0.25 & 0.93 & 0.21 & 0.21 & 0.19 \\
    Llama3.1-8b & 0.70 & 0.04 & 0.74 & 0.68 & 0.05 & 0.72 & 0.04 & 0.04 & 0.04 \\
    Llama3.1-70b & 0.89 & 0.28 & 0.89 & 0.88 & 0.40 & 0.89 & 0.26 & 0.37 & 0.26 \\\midrule 
    GPT-3.5-turbo & 0.84 & 0.35 & 0.93 & 0.81 & 0.37 & 0.84 & 0.32 & 0.28 & 0.28 \\
    GPT-4o-mini & 0.96 & 0.35 & 0.93 & 0.93 & 0.37 & 0.96 & 0.30 & 0.35 & 0.30 \\
    GPT-4o & \textbf{0.98} & \textbf{0.44} & 0.93 & \textbf{0.93} & \textbf{0.42} & \textbf{0.98} & \textbf{0.39} & \textbf{0.42} & \textbf{0.39} \\
    \bottomrule
    \end{tabular}

%% file: data/table/judge.tex

\begin{table*}[h!tbp]
    \caption{Judging results.}
    \centering

    \begin{subtable}{0.78\textwidth}
    \caption{Detailed f0.5 scores. We show the SAG score and in parenthesis the GAG score. Legend: AVGO: (resp. AVGS) average f0.5 over all criteria when judging other models' (resp. the judge's own) feedback.}
    \resizebox{0.99\textwidth}{!}{
    \begin{tabular}{l|ccc|ccc|cc}
\toprule
judge & EA & ES & EC & FA & FS & FC & AVGO & AVGS \\
\midrule
Phi-3-mini & 0.70 (+0.00) & 0.41 (-0.03) & 0.81 (-0.02) & 0.65 (+0.00) & 0.37 (-0.04) & 0.78 (+0.00) & 0.58 (-0.02) & 0.41 (+0.09) \\
Mistral & 0.67 (+0.07) & 0.14 (+0.50) & 0.81 (-0.02) & 0.63 (+0.05) & 0.19 (+0.42) & \textbf{0.81} (-0.04) & 0.49 (+0.18) & 0.44 (+0.13) \\
Llama-3.1-8b & 0.63 (+0.12) & 0.55 (+0.15) & 0.75 (+0.04) & 0.63 (+0.12) & 0.56 (+0.09) & 0.75 (+0.04) & 0.60 (+0.11) & 0.51 (+0.22) \\
Ensemble & 0.68 (+0.06) & 0.28 (+0.33) & 0.81 (-0.01) & 0.64 (+0.06) & 0.33 (+0.24) & 0.79 (+0.01) & 0.53 (+0.14) & / \\
Llama-3.1-70b & 0.69 (+0.17) & 0.72 (+0.10) & \textbf{0.81 (+0.01)} & 0.71 (+0.10) & 0.71 (+0.10) & 0.80 \textbf{(+0.03)} & \textbf{0.69 (+0.14)}  & \textbf{0.69} (+0.14) \\ \midrule
GPT-3.5-turbo & 0.65 (+0.00) & 0.09 (+0.19) & 0.79 (-0.01) & 0.65 (+0.01) & 0.20 (+0.26) & 0.78 (-0.01) & 0.46 (+0.09) & 0.51 (+0.08) \\
GPT-4o-mini & \textbf{0.74} (+0.08) & 0.66 (+0.16) & 0.75 (+0.01) & \textbf{0.74} (+0.05) & 0.70 (+0.12) & 0.76 (+0.02) & 0.64 (+0.13) & 0.61 (+0.24) \\
GPT-4o & 0.72 (\textbf{+0.16}) & \textbf{0.78 (+0.09)}  & 0.74 (+0.04) & 0.72 \textbf{(+0.12)} & \textbf{0.76 (+0.11)} & 0.77 (+0.02) & 0.68 (+0.12) & 0.66 \textbf{(+0.26)} \\
\bottomrule
\end{tabular}
        
    }
    \end{subtable}
    \hfill
    \begin{subtable}{0.19\textwidth}
    \centering
    \caption{Kappa scores. \\ SAG (+- GAG diff)}
    \resizebox{\textwidth}{!}{
    \begin{tabular}{lc}
    \toprule
    judge & kappa \\
    \midrule
    Phi-3-mini & 0.09 (+0.00) \\
    Mistral & 0.03 (+0.23) \\
    Llama-3.1-8b & 0.12 (+0.27) \\
    Ensemble & 0.06 (+0.20) \\
    Llama-3.1-70b & \textbf{0.36 (+0.29)} \\ \midrule 
    GPT-3.5-turbo & 0.02 (+0.08) \\
    GPT-4o-mini & 0.25 (+0.27) \\
    GPT-4o & 0.34 (+0.26) \\
    \bottomrule
    \end{tabular}

    }
    \end{subtable}
    \label{tab:judge_results}
\end{table*}

%% file: sections/05_discussion.tex
\section{Discussion}

\paragraph{Teaching and Learning Implications.}
While the educational community has been focused on leveraging proprietary models, our study aims to show that alternative solutions are accessible to educators and practitioners. In particular, Llama-3.1-70B performs on par with GPT-3.5-turbo for feedback generation and competes with GPT-4o for judging the quality of other language models generated feedback. Educators could leverage such a model in building their feedback tools or for benchmarking purposes \cite{koutcheme-etal-2024-using,phung2023generative}.
Notably, the size of a language model no longer correlates directly with performance. For example, a smaller language model like the Phi-3-mini competes with the 7B Mistral and 8B Llama models, offering promising feedback generation and judging performance. 
While this model (like multiple of the others) struggles with hallucination, we believe that fine-tuning techniques might alleviate this issue, and strengthen the results~\cite{kotalwar2024hintsinbrowserbenchmarkinglanguagemodels}.

\paragraph{Practical deployment.}
Importantly, these open-source models are also easy to access thanks to APIs offered by companies such as \textsc{HuggingFace}. For instance, for conducting our experiments, using the \textsc{Transformers} library, all models were freely accessible, except for Llama-3.1-70B (which required paying 9 dollars for a month of rate-limited access). We acknowledge that using an external API to query open-source language models might defeat the purpose of data privacy. However, several institutions leverage ChatGPT APIs in one way or another \cite{rongxin_teaching_2024}, and HuggingFace platforms, which are dedicated to open-source, offer the same data privacy guarantees. Moreover, prior works show evidence that smaller models (such as Phi-3-mini) could be deployed on consumer devices \cite{liu_2024_can}, or even in browsers \cite{kotalwar2024hintsinbrowserbenchmarkinglanguagemodels}, alleviating the need to rely on external API. 

\paragraph{Limitations.}
Our work has limitations. The prompts we used likely influenced the results, and more specific prompts or alternative prompting strategies (which we did not explicitly compare) could impact model performance. Also, we only considered introductory programming assignments written in Python and not other programming languages. Moreover, we only considered two types of feedback, but other types such as hints, (with potentially more pedagogical benefits) exist. Our selection of language models, although considered the recent state-of-the-art at the time of writing, does not exhaust all possible alternatives to popular models. Our labelling process is also not perfect, as we only used two raters, which resulted in a moderate inter-rater reliability score (0.54), and our grading criteria did not use actual students (the intended audience) to rate the clarity of the outputs. Additionally, for judging feedback, we could have used judge language models which are specifically designed for evaluation, but we used generic language models instead. Finally, our report lacks a discussion (and examples) of the specific type of issues encountered when generating and judging feedback.

\paragraph{Future Work.}
In the future, we will conduct a larger-scale evaluation of open-source language models' ability to generate (and judge) other types of feedback and support. In particular, we are extending the Socratic benchmark to include ground truth next-step hints~\cite{roest2024next}, and we are running an evaluation of language models' ability to generate such hints \cite{kotalwar2024hintsinbrowserbenchmarkinglanguagemodels}, as well as their ability for being Socratic guides. 
Beyond tracking models' performances, we aim to improve the ability of small language models (e.g. Phi-3-mini) to be teaching assistants by using Reinforcement Learning techniques to tackle the selectivity problem \cite{scarlatos_improving_2024}. The varied strengths of different models also suggest that a combined approach (ensemble methods) might yield even better results for feedback generation. 
As our LLM jury results contrasted those by Verga et al.~\cite{verga2024replacingjudgesjuriesevaluating}, we also intend to conduct a study on how the variability, individual performance, and the number of judges in the LLM jury affect judging performance. 

%% file: sections/06_conclusions.tex
\section{Conclusions}

In this paper, we evaluated (1) how language models perform in providing explanations of issues in programs and generating bug fixes (RQ1), and (2) how well different language models, including open-source ones, perform in evaluating the quality of feedback generated by other LMs (RQ2).
Our paper highlights that top open-source language models are valid competitors to proprietary language models for both generating and assessing the quality of programming feedback. Open-source language models could provide benefits for powering free tools, which is particularly important for institutions with limited funding.
As an additional contribution, we release the code used to conduct our experiments, including the models' outputs and the annotators' responses\footnote{\url{https://github.com/KoutchemeCharles/feed_genju}}.

%% file: sections/07_acknowledgements.tex
\begin{acks}
   This research was partially supported by the Research Council of Finland (Academy Research Fellow grant number 356114). 
\end{acks}